\documentclass[11pt]{article}

\usepackage{authblk}
\usepackage[T1]{fontenc}
\usepackage{lmodern}
\usepackage[margin=1in]{geometry}
\usepackage{setspace}

\usepackage{tabularx}
\newcolumntype{Y}{>{\raggedright\arraybackslash}X}

\usepackage{natbib}
\setcitestyle{authoryear,round}
\usepackage{hyperref}
\hypersetup{hidelinks}

\title{From More to Better: Causal Inference and Digital Experimentation in the Age of AI}

\author[1]{Winston Chou}
\author[2]{M{\aa}rten Schultzberg}
\author[3]{Lo-Hua Yuan}
\author[4]{Wenjing Zheng}
\author[5]{Avi Feller}
\author[6,9]{David Holtz}
\author[7]{Widad Machmouchi}
\author[6]{Sven Schmit}
\author[8]{Gagan Bhat}
\author[1]{Simon Ejdemyr}
\author[1]{Veronica Hannan}
\author[7]{Amy Hartwig}
\author[10]{Mahnaz Islam}
\author[1,12]{Nathan Kallus}
\author[4]{Jeremy Karp}
\author[2]{Kyle Kretschman}
\author[9]{Hannah Li}
\author[10]{James McQueen}
\author[8]{Evan Miller}
\author[11]{Nicolas Stier}
\author[6]{Sean Taylor}

\affil[1]{Netflix}
\affil[2]{Spotify}
\affil[3]{Independent}
\affil[4]{Roblox}
\affil[5]{University of California, Berkeley}
\affil[6]{OpenAI}
\affil[7]{Microsoft}
\affil[8]{Anthropic}
\affil[9]{Columbia University}
\affil[10]{Amazon}
\affil[11]{Meta}
\affil[12]{Cornell University}

\date{}

\begin{document}

\maketitle

\begin{abstract}
Artificial intelligence (AI) is quickly expanding the capacity of organizations to produce and analyze data. We argue that this expansion shifts the leverage of data scientists from performing individual analyses to assessing the credibility and importance of findings and determining when evidence generalizes across contexts. Drawing on a roundtable of academic researchers and industry practitioners, we examine three ways in which data scientists who work on causal inference and experimentation can help organizations turn more analytical capacity into better decisions: preserving and accumulating the learning value of experimentation beyond individual tests; choosing and combining experimental, observational, and synthetic evidence; and ensuring evaluation keeps pace with AI as its capabilities and uses evolve. In each area, we distinguish established practice from open questions. We conclude with implications for the training and practice of data science and for organizations.
\end{abstract}

\doublespacing

\section{Introduction}

Artificial intelligence (AI) is transforming how organizations generate, analyze, and use data. As the barriers to writing code and analyzing data fall, more people can investigate questions that previously required specialized expertise. Yet more analytical capacity does not necessarily yield deeper understanding or better decisions. The productivity gains from earlier general-purpose technologies, such as electricity and the internet, depended on complementary if intangible investments in processes and skills \citep{brynjolfsson2021productivity}. Likewise, the value of more analysis will depend on what organizations build around it.

We argue that, as AI makes data and analyses easier to generate, the central challenge in causal inference and experimentation is shifting away from producing individual analyses to assessing the relevance and credibility of findings and determining when they generalize across contexts. Observational causal inference offers a useful analogy: an observational estimate, even when backed by millions of data points, is credible only under assumptions that link the data to the causal question and counterfactuals at hand \citep{hernan2006}. Similarly, AI equips organizations with exciting new capabilities, such as the simulation of new treatments on synthetic respondents; yet whether and how these capabilities should inform decisions affecting real users requires assumptions about internal validity, external validity, and transportability that should be stated and scrutinized \citep{pearl2014,degtiar2023,egami2023,persson2026}.

To be sure, causal inference and experimentation have always involved more than estimating treatment effects. Understanding why an intervention works, determining whether its effects generalize, reconciling conflicting findings, and connecting empirical results to meaningful outcomes are longstanding challenges in digital experimentation \citep{gupta2019, holtz2026}.  However, AI intensifies these challenges in three distinct ways. By accelerating product development and lowering barriers to analysis, AI increases both the demand for measurement and the volume of results that must be critically evaluated. As a source of synthetic evidence, AI expands what organizations can simulate without testing on real users — though, again, drawing the appropriate conclusions from these simulations requires causal assumptions linking simulated responses to real user behavior \citep{persson2026}. Lastly, as an open-ended and rapidly evolving product, AI upends static measures of success, especially as users discover new applications and adapt their behavior \citep{park2026}.

In this paper, we outline three ways data scientists can help organizations meet these challenges:
\begin{enumerate}
    \item \textbf{Understand causal mechanisms and build organizational knowledge.} As AI accelerates product development and analysis, data scientists can help organizations move beyond individual experiments and decisions toward cumulative learning. We examine how causal reasoning and experimental metadata can help produce and preserve understanding of generalizable mechanisms that explain whether interventions succeed or fail.
    \item \textbf{Establish playbooks for how to choose and combine different sources of evidence to support a decision.} As synthetic, observational, experimental, and hybrid sources of evidence become more accessible, organizations need to be aware of their assumptions and limitations. We discuss the considerations behind choosing and combining these sources and where the validity of synthetic evidence remains unresolved.
    \item \textbf{Build evaluation frameworks that adapt to evolving AI-powered products.} As products become more open-ended and their behavior less predictable, organizations must actively evolve what they measure and scrutinize whether and how past findings generalize. We examine what existing methods can establish and where novel evaluation frameworks may be needed.
\end{enumerate}

Our perspective draws on the 2026 Causal Inference and Digital Experimentation Roundtable (CIDER), which convened participants from technology companies and academic institutions to discuss the intersection of data science and AI.\footnote{CIDER consists of an around-the-table shareout of each organization’s top challenges, a vote on which challenges to elevate, and unscripted discussion of the topics that received the most votes. For a broad overview of open challenges in causal inference and experimentation, we summarize the challenges raised across participating organizations in the Appendix.} From the CIDER discussions and the authors' collective experience, we distill several key claims, highlighted in \textbf{bold}. We use the following vocabulary to communicate the level of agreement among authors. \emph{Consensus} refers to propositions broadly endorsed by the author group. \emph{Shared} refers to propositions expressed by multiple authors without substantial pushback. \emph{Contested} refers to propositions having substantive disagreement between authors.

\section{From Individual Experiments to Organizational Learning}

As AI accelerates feature development, organizations face growing pressure to evaluate more ideas. Yet when experiments are conducted, organizations often fail to capture their learning value beyond the immediate decision (such as the choice to ``launch'' a tested innovation). An experiment generates learning value when it changes a relevant belief or informs a subsequent action. An experiment can fail to justify the launch of a new feature or the scaling up of a novel intervention and still teach the organization something useful. However, without recording the experimenter's intent, observations, and reasoning, this learning value can be difficult to share and reconstruct. 

\subsection{Mechanisms and metadata for organizational learning}

Our first proposition is that, as AI increases the volume of experimentation, \textbf{data scientists should prioritize learning across experiments}. Of special importance are generalizable mechanisms that help explain why interventions succeed or fail. Mechanism metrics --- variables hypothesized to mediate the effects of treatment on downstream outcomes --- are key to diagnosing whether an intervention was delivered as intended, whether the intermediate processes it sought to affect changed, and whether relevant user or business outcomes improved. Mechanistic reasoning can also guide follow-up experiments, including tests that probe the mechanism directly.\footnote{While mechanisms are essential to explaining experimental outcomes, establishing causal mediation is difficult without additional assumptions or specialized experimental designs \citep{imai2010}. Correlation between estimated treatment effects on a mechanism metric and a business outcome does not establish that changes in the mechanism caused changes in the business outcome: other causal pathways or correlated estimation error may explain the relationship \citep{bibaut2024}. For similar reasons, using a mechanism metric as a surrogate for downstream business outcomes requires validation and care \citep{prentice1989}.}

A corollary of this proposition is that learning and launching are distinct goals. The launch decision asks whether a treatment should be released to users or whether an intervention should be scaled up. The learning objective asks whether the experiment improved the organization's understanding of users, products, constraints, and tradeoffs. These goals can diverge: an experiment that fails to justify a change can nevertheless rule out a hypothesized mechanism or inform which interventions a team pursues next. Some tests are designed as learning tests that will not launch regardless of the results. For example, at Netflix, experiments are run to test structural models \citep{zielnicki2026} and to measure the cumulative impact of an experimentation program \citep{aridor2026}.

Accumulating knowledge across experiments necessitates records of what was tested and how the results were interpreted, not just estimated treatment effects. To interpret a past result and assess its relevance to a new decision, teams need to know what experimenters tried to test; which core, mechanism, and guardrail metrics actually changed; which users were affected; and how the findings informed subsequent decisions. Code and feature flags can help verify what was actually implemented. Circumstances such as the user population, product state, and model version can also affect whether a past result carries over to a new decision. Records should distinguish statistical findings from proposed explanations and include inconclusive, abandoned, and screened-out ideas. Otherwise, a repository may disproportionately preserve successes and memorable failures. Especially as experimentation scales, incomplete metadata exacerbates the risk of repeatedly testing similar ideas without accumulating understanding \citep{kohavi2020}.

In many organizations, this institutional knowledge is often tacit or scattered across documents, if written down at all. One promising application of AI is to extract structure from code changes, strategy documents, and other unstructured artifacts. These records, especially when reviewed by human experts and traceable to authoritative sources, can also provide vetted context for AI agents that draw on past experiments to propose or evaluate new interventions.

\subsection{Balancing structure and burden in metadata collection}

The value of experimental metadata was consensus; how much structure to require in collecting it was contested. Some authors advocated for structured elicitation of metadata, for example, requiring hypotheses, causal diagrams, or intended mechanisms before running a test. Others noted that requiring such structure can result in false precision and/or low-quality responses from annoyed partners. Furthermore, complex hypotheses may be difficult to represent adequately in a structured manner. One compromise is to ask for lightweight structure in exchange for automation: for example, specify the intended treatment(s) and primary metrics for sample size calculation, or record launch decisions to support cumulative impact measurement.

\section{Beyond Classical A/B Testing: Synthetic and Hybrid Measurement}

How organizations choose and combine evidence should depend on the question or decision at hand. This principle applies equally to experimentation and synthetic or observational evidence. For example, even a well-designed fixed-allocation A/B test may be inappropriate when adaptive experimentation would better serve the decision \citep{hadad2021}, or when faster and less costly sources of evidence would suffice. AI expands the set of measurement options available to organizations, increasing the importance of deciding which sources of evidence are appropriate for a given decision.

\subsection{Evaluating and combining diverse sources of evidence}

Our core proposition here is that \textbf{evidence must be validated against its intended use}. A model used to screen ideas need not estimate their effects precisely, but it should reliably distinguish promising candidates from those unlikely to succeed. In contrast, a model used to support launch decisions should reliably estimate the magnitude of user benefits and harms.

Experimentation remains the gold standard for evidence of real user behavior when exposed to novel interventions. However, organizations overuse it when they treat it as the sole arbiter of decisions requiring multiple sources of evidence \citep{deaton2018}. A useful metaphor is the ``evidence ladder'': synthetic and offline evaluations are useful for screening ideas, pilots and online experiments for assessing promising candidates, and longer-term holdbacks for measuring long-term cumulative impact. Live experiments remain important for evaluating effects on real users, but not every decision needs to ascend the entire ladder.

Large language models (LLMs) may reproduce observed human responses, but this does not establish that they can estimate differences between treatments or predict the effects of new interventions. Existing research on the fidelity of synthetic respondents paints a mixed picture of their capabilities. \citet{argyle2023} and \citet{horton2023} find that LLMs are able to reproduce aspects of human behavior in political surveys and economic experiments, respectively. These results provide evidence of usefulness in particular settings, but leave open how that performance might translate to new tasks or interventions.

\citet{toubia2025} introduce Twin-2K-500, a dataset for building and evaluating personalized synthetic respondents (``digital twins''). Although their digital twins predict held-out individual responses better than a random benchmark, they reproduce only some of the experimental patterns observed among human participants. \citet{gao2025} find that the models they study generally fail to reproduce human choices and responses to changes in a strategic decision task. \citet{wang2025} find that LLMs can flatten variation within demographic groups and sometimes resemble outsiders' portrayals more closely than group members' own responses. 

If recent history is a guide, the capabilities of LLMs to emulate human behavior are likely to improve over time. However, the underlying principle --- to evaluate models against their intended use --- remains the same.  For example, models that screen new treatments should be audited on a random subset of the ideas they reject (and which are otherwise suitable for live user testing). Models used to predict treatment effects should be evaluated against held-out experiments, with special care to ensure that the experimental results do not leak into their training data.

Combining multiple sources of evidence is a well-worn strategy that can help organizations avoid making sweeping judgments about different types of data. Prediction-powered inference, for example, combines many model predictions with a smaller sample of measured outcomes. Comparing the two allows researchers to correct the final estimates for prediction error while accounting for uncertainty in the correction \citep{angelopoulos2023}. The surrogate index is another example of data combination: it uses observational data to relate short-term proxies to a long-term outcome, then applies that relationship to short-term outcomes measured in an experiment. Under surrogacy and comparability assumptions, this allows researchers to estimate long-term treatment effects before those outcomes can be observed in the experiment \citep{athey2026}.

Crucially, these methods are valid only under assumptions that can be clearly articulated and scrutinized. Combining human and synthetic responses requires similar care: researchers need to specify how synthetic responses relate to human responses and how prediction errors affect the resulting estimates and uncertainty \citep{persson2026}.

\subsection{When can synthetic evidence substitute for live experiments?}

Although the need to validate evidence against its intended use was a widely shared proposition, whether synthetic evidence can replace live experiments remains contested. The case for synthetic evidence is strongest when live user data is expensive, slow, scarce, or risky to collect. But the same conditions that make simulation attractive can make bias costly. Self-driving is a high-stakes example: simulation is vital for reducing exposure to risk, but failure to simulate rare real-world scenarios can be catastrophic \citep{schwall2020,scanlon2021}.

To be sure, live user experiments also have their limits. Detecting rare outcomes or estimating effects across interacting users requires an appropriate experimental design and sufficient observations \citep{hanley1983}. These limitations strengthen the case for considering and combining experiments with other evidence. Existing statistical concepts such as surrogacy and transportability provide a vocabulary for making some of the necessary assumptions explicit \citep{persson2026}. Organizations should also ask the practical question of what evidence from live users would change a decision, especially if that evidence can itself be noisy or nondefinitive.

\section{Evaluating Non-Deterministic and AI-Powered Systems}

Task ambiguity, user learning and adaptation, and rare failures are longstanding challenges in the evaluation of digital products. AI products intensify these challenges through the sheer variability of their applications and outputs. A key tension here is that, although organizations must continuously revise their evaluations as these use cases emerge, they also need to curate an ever-expanding set of evaluations and distill the measurements that are most consequential for decision-making and product development. We encapsulate this tension with the next two propositions.

\subsection{Evaluating products as their uses change}

Our first proposition here is that \textbf{evaluation needs evolve with how a product is used}. This proposition may seem self-evident, yet the challenge of keeping pace with AI products is not always fully appreciated. For example, a two-week experiment on a coding agent might cover routine debugging, whereas after deployment users adopt it for code migrations. Under appropriate assumptions, reweighting can account for differences in the prevalence of tasks represented in the experiment, but it cannot recover effects for entirely unrepresented tasks, regardless of how much data was collected in the experiment \citep{dahabreh2019}. Even when tasks remain similar, how users interact with the product can change \citep{sigerson2026}. Some users might become more adept at prompting the model for certain tasks. Others deterred by an early bad experience might not return, even after the model improves.

Although the authors shared the view that AI products present unique challenges, whether these challenges warrant genuinely new methodologies and evaluation frameworks was contested. On the one hand, tools such as controlled rollouts, guardrail metrics, and anytime-valid inference remain useful for catching unexpected regressions and undesirable behavior \citep{howard2021, johari2022}. Counterfactual or ``shadow'' logging, widely used in advertising measurement \citep{johnson2017ghost}, offers another useful tool: recording what an alternative system would have done without exposing users to its output. For AI products, shadow evaluations can run candidate and incumbent models on the same live inputs while showing users only the incumbent's response. This enables paired comparisons of output quality, safety, and computational cost \citep{machmouchi2023evaluate}.

On the other hand, the multimodal and qualitative nature of LLM output increases the need for new and correspondingly rich evidence sources. For example, screenshots, user reports, and human review can all reveal failures that aggregate measures miss \citep[cf.][]{shankar2024}.

\subsection{Connecting evaluations to product decisions}

The previous proposition sits in tension with our next proposition, which is that \textbf{each metric and evaluation should have a clear role in decision-making}. Continuously adding evaluations for new tasks and failures is important for maintaining coverage of how the product is used. However, each new evaluation also makes it harder to interpret the results holistically.

Not every evaluation needs to determine whether a model should launch. For example, an evaluation of a particular debugging task can help developers identify and fix a weakness of a coding agent, but whether that weakness should block deployment depends on how often users encounter it and how serious its consequences are. Other evaluations serve as guardrails: rare but damaging model failures might be sufficient to block a launch even when average quality improves. In general, each measure should have a clear role in decision-making based on its relationship to higher-level organizational goals \citep{rodden2010}.

How exactly to tame the proliferation of metrics and evaluations was contested. Some authors focused on the infrastructure needed to run, maintain, and inspect large numbers of evaluations. Others focused on measuring the connection of each metric to organizational north stars, such as long-term user retention and revenue. Both concerns bear on the question of which measurements to add, update, or retire. Data scientists can help teams make these choices by specifying what each evaluation is intended to establish and how its results should affect product development or deployment.

\section{Implications for Data Science Practice and Organizations}

\subsection{Data science practice and training}

As models become more adept at analyzing data, we anticipate that data scientists will spend more time selecting and framing important questions, evaluating the systems and workflows responsible for producing analyses, interpreting results, and making decisions. Foundational and frontier expertise are complementary here. On the one hand, foundational expertise in statistical, computational, and domain knowledge can help with specifying an agent's task, recognizing analytical errors, and evaluating its conclusions \citep{nasem2018}. On the other hand, hands-on experience with AI tools can help with understanding their capabilities and failure modes.

A shared concern is that, although the authors generally expect the importance of context, judgment, and communication skills to increase, AI may automate some of the tasks through which junior data scientists have traditionally acquired those assets. For example, using AI, an analyst can quickly generate a working query without understanding the underlying data model or intuiting when the result may be misleading. In a relevant study of developers learning an unfamiliar programming library, \citet{shen2026} found that AI assistance impaired conceptual understanding, code reading, and debugging abilities. Providing forums where humans review AI-generated analyses or consequential decisions together is one way that organizations can help junior data scientists hone their data literacy and judgment.


\subsection{Organizing analytical work}

Consider two agents that produce equally sophisticated analyses of the same experiment using different populations or observation windows. Both analyses may be correct, but address subtly different questions. The organization needs to determine which question most closely corresponds to the decision at hand. Ideally, it should also record the rationale behind its choice, so that even this decision --- how an organization chooses to learn --- can be audited and improved over time.

Some of this work can be incorporated into analytical infrastructure \citep{cherry2026}. Standard calculations and filters can be implemented as deterministic routines for agents to invoke and assemble. More nuanced choices, such as navigating metric tradeoffs and reconciling conflicting findings, may require explicit organizational alignment and deliberate curation. Mechanisms for designating ``blessed'' analyses, such as watermarks, can help organizations keep a clean record of findings and decisions, again to support organizational learning.

How exactly analytical work will be organized remains contested. Some authors expect analyses to flow through more centralized infrastructure, and for scientific work to index more heavily on building and evaluating those analytical systems. Others expect decentralization, as seasoned analysts can investigate more questions with greater nuance and depth without falling back to one-size-fits-all tooling. Yet another possibility lies somewhere in between, with organizations supplying shared context and primitives (e.g., metric definitions) to support arbitrarily deep analyses.

\section{Conclusion}

AI is expanding the capacity of organizations to generate ideas, data, and analyses. Data scientists can equip organizations to harness this capacity by preserving and accumulating knowledge across experiments and over time, choosing and assembling the right mix of evidence for a given decision, and revising evaluations as products and their uses evolve.

Important questions remain. What metadata is sufficient for organizations to accumulate knowledge, and how do we make its collection incentive-compatible? How reliable is synthetic evidence as a substitute for live user traffic, and how can the two be effectively combined? When should an evaluation be added or retired, and what evidence connects it to long-term value? Given the rapid pace of AI development, the more technical of these questions may well be resolved by the next CIDER. However, other questions --- especially those that stem from the underlying challenges of helping organizations learn, innovate, and make decisions under uncertainty --- are likely to remain evergreen.

\bibliographystyle{plainnat}
\bibliography{cider_paper}

\appendix
\section{Challenges identified at CIDER 2026}

Each CIDER begins with an around-the-table shareout of the top experimentation and causal inference challenges confronting each organization. This year's discussion surfaced six broad clusters, summarized in Table~\ref{tab:clusters}: evolving measurement needs, increasing business complexity, deploying and evaluating data science agents, experimentation velocity, alternatives and complements to the classical A/B test, and evaluation of AI-driven products.

\begin{table}[htbp]
\centering
\caption{Challenge clusters shared across organizations that participated in CIDER 2026}
\label{tab:clusters}
\small
\begin{tabularx}{\textwidth}{|>{\raggedright\arraybackslash}p{1.5in}|Y|}
\hline
\textbf{Cluster} & \textbf{Summary}\\
\hline
Measurement &
As products mature, core metrics like retention and Daily Active Users (DAU) are harder to move and user satisfaction is more nuanced as user journeys get longer. Teams need sensitive proxies that capture user value and that feature teams can ``hill climb'' on. Building such metrics is an acute challenge, especially for AI-powered products: for example, longer coding assistant sessions could reflect delight or frustration, hence the need for metrics that can distinguish between these two scenarios.\\
\hline
Business complexity &
Experimentation leaders are increasingly asked to help navigate tradeoffs between user value, business value, engagement, retention, and infrastructure costs. Local wins can create global losses. Comparing disparate outcomes, measuring cumulative effects, and aligning incentives across teams all remain difficult. Domain knowledge and product experience continue to be critical for such decisions, and they are increasingly harder to operationalize for complex products.\\
\hline
Data science agents &
Agents are both tools and objects of evaluation. They often need institutional context---shared metric definitions, default filters, and approved sources---to perform well \citep{cherry2026}. The democratization of experimentation analysis through AI has empowered teams to understand the impact of changes faster. At the same time, it amplifies governance risk and the need for certification of ``blessed'' analyses to avoid untrustworthy decisions based on inaccurate interpretation of data.\\
\hline
Experimentation velocity &
AI lowers the cost of generating and testing ideas, shortening the path between ``prompt to prototype to production.'' Because live user traffic and manual review capacity remain scarce, filtering ideas becomes critical; AI funnels need their own validation. Higher throughput increases the value of historical learning: reusable metrics, portfolio-level views, and records of what has already been tried.\\
\hline
Alternatives and complements to classical A/B testing &
Options include incremental rollouts, holdbacks, observational causal inference, adaptive experiments, bandits, offline simulations, synthetic users, and hybrid systems. The right combination depends on the decision and the evidence each method can provide. This also amplifies the need to build guardrails and harnesses around such methods as AI makes them increasingly available.\\
\hline
Evaluation of AI-powered products &
AI products can support open-ended tasks and exhibit branching and nondeterministic behavior. This complicates metric selection and the detection of rare but costly failures. Teams must reconcile offline benchmarks, online experiments, qualitative and anecdotal evidence (e.g., screenshots), and continuous monitoring. Using AI agents for evaluation requires additional care to avoid feedback loops where AI agents evaluate themselves.\\
\hline
\end{tabularx}
\end{table}

\end{document}